\documentclass{article}

\usepackage{arxiv}

\usepackage[utf8]{inputenc} 
\usepackage[T1]{fontenc}    
\usepackage{url}            
\usepackage{booktabs}       
\usepackage{array}
\usepackage{amsfonts}       
\usepackage{amsmath}
\usepackage{bm}
\usepackage{nicefrac}       
\usepackage{microtype}      
\usepackage{lipsum}		
\usepackage{graphicx}
\usepackage{subfig}
\graphicspath{images}
\usepackage{natbib}
\usepackage{doi}
\usepackage{siunitx}
\usepackage{booktabs}
\usepackage{multirow}

\usepackage{xcolor}
\definecolor{dgreen}{RGB}{0, 125, 0}
\definecolor{dred}{RGB}{200, 0, 0}
\definecolor{dblue}{RGB}{0, 0, 200}

\usepackage{hyperref}
\hypersetup{
	colorlinks=true,
	linkcolor=dred,
	filecolor=magenta,      
	urlcolor=dblue,
	citecolor=dgreen
}

\newcommand\freefootnote[1]{%
  \let\thefootnote\relax%
  \footnotetext{#1}%
  \let\thefootnote\svthefootnote%
}
\usepackage[hang,flushmargin]{footmisc}

\usepackage{amssymb}
\usepackage{pifont}
\newcommand{\xmark}{\ding{55}}%
\usepackage{floatrow}
\newfloatcommand{capbtabbox}{table}[][\FBwidth]

\title{Multi-Modal Learning-based Reconstruction of High-Resolution Spatial Wind Speed Fields}

\author{ Matteo Zambra \quad Nicolas Farrugia \quad Ronan Fablet \\
	IMT Atlantique, URM CNRS Lab-STICC\\
	29280 Brest, France \\
	\texttt{\{matteo.zambra, nicolas.farrugia, ronan.fablet\}@imt-atlantique.fr} \\
	\And
	Dorian Cazau \\
	ENSTA Bretagne, URM CNRS Lab-STICC\\
	29200 Brest, France \\
	\texttt{dorian.cazau@ensta-bretagne.fr} \\
	\And
	Alexandre Gensse \\
    Naval Group \\
    83090 Toulon, France \\
    \texttt{alexandre.gensse@naval-group.com}
	}
\date{}

\renewcommand{\shorttitle}{}

\hypersetup{
pdftitle={zambra-et-al-2022},
pdfsubject={physics.comp-ph, physics.ao-ph, cs.LG},
pdfauthor={Matteo Zambra},
pdfkeywords={variational data assimilation, deep learning, geophysical dynamics},
}

\begin{document}
\maketitle

\begin{abstract}
Wind speed at sea surface is a key quantity for a variety of scientific applications and human activities. Due to the non-linearity of the phenomenon, a complete description of such variable is made infeasible on both the small scale and large spatial extents. Methods relying on Data Assimilation techniques, despite being the state-of-the-art for Numerical Weather Prediction, can not provide the reconstructions with a spatial resolution that can compete with satellite imagery. In this work we propose a framework based on Variational Data Assimilation and Deep Learning concepts. This framework is applied to recover rich-in-time, high-resolution information on sea surface wind speed. We design our experiments using synthetic wind data and different sampling schemes for high-resolution and low-resolution versions of original data to emulate the real-world scenario of spatio-temporally heterogeneous observations. Extensive numerical experiments are performed to assess systematically the impact of low and high-resolution wind fields and in-situ observations on the model reconstruction performance. We show that in-situ observations with richer temporal resolution represent an added value in terms of the model reconstruction performance. We show how a multi-modal approach, that explicitly informs the model about the heterogeneity of the available observations, can improve the reconstruction task by exploiting the complementary information in spatial and local point-wise data. To conclude, we propose an analysis to test the robustness of the chosen framework against phase delay and amplitude biases in low-resolution data and against interruptions of in-situ observations supply at evaluation time.\freefootnote{\small  This work is to be submitted to the IEEE for possible publication. Copyright may be transferred without notice, after which this version may no longer be accessible.}
\end{abstract}



\section{Introduction}\label{sec:introduction}
Despite its operational and scientific importance, wind speed at sea surface is not fully describable at both large extent and smaller scales. 
Among the diverse sources of sea-surface information available, the state-of-the-art for wind speed forecasting and hindcasting is represented by Data Assimilation-based products, called \emph{reanalyses} (\cite{Storto2019, Valmassoi2023current}. These products are a fundamental knowledge base for climate research and meteorological applications. Reanalyses may have hourly temporal resolution but their spatial resolution may not satisfy operational needs.
An alternative is represented by remote sensing techniques, such as satellite imaging. In recent years, Synthetic Aperture Radar (SAR, see \cite{RANA2019, li2020}) imagery has gained particular attention. One major limitation in satellite remote sensing, especially with SAR imaging, is the temporal scarcity of data on a same region, since the time interval between two consecutive observations of the same zone can be from 12 hours up to some days. This limitation prevents a dynamical characterization of wind speed on a spatially wide region. A third source of useful information is represented by in-situ observations (\cite{Gould2013situ}). In-situ data are reliable measurements of the atmospheric state. These observations, thanks to their sub-hourly measurement capabilities, can sense the phenomenon near-continuously but they are local in space and lack a large spatial coverage.

Our objective is to retrieve the complete high-resolution dynamical information of sea-surface wind speed using its partial observations, namely spatial and local high-resolution observations and low-resolution NWP products. The consolidated approach in the geophysical community to solve this kind of \emph{inverse problems} is to use Data Assimilation (see~\cite{carrassi2018data-assimilation} and~\cite{bannister2017}). Recently, schemes that bridge Data Assimilation and Deep Learning approaches have gained popularity (\cite{Cheng2023, Arcucci2021deep, Bonavita2021machine}). Deep Learning can be used as a complement in Data Assimilation schemes, for example to improve the background state resolution (\cite{Barthélémy2022}) or to represent model errors (\cite{Farchi2022model}). Other examples of Data Assimilation and Deep Learning interactions involve end-to-end learning systems, that implicitly represent the system physical dynamics (\cite{fablet2021jointinterpolation, beauchamp2020, fablet2021endtoend}). The latter case is particularly attracting since deep learning computational efficiency and representation power are fully exploited. In particular, in this work we propose the application of the 4DVarNet scheme (\cite{fablet2021learning-james}) for the solution of this inverse problem. Its appeal stems from the complete parameterization of the underlying state-space dynamics using trainable neural networks. In addition, the 4DVarNet framework proposes a trainable neural gradient solver designed to speed up the model convergence. Probably, the most valuable aspect of this model is the dynamical characterization of the system state, as the 4DVarNet, consistently with the 4DVar Data Assimilation scheme (\cite{Bannister2001elementary, Talagrand2015}), relies on a state-space formulation of the system evolution. 

We run the experiments of synthetic model data, which are processed to simulate the different data sources mentioned above. The main contribution of this work may be summarized as follows. First, we set up an Observation System Simulation Experiment framework (\cite{Hoffman2016future}) to assess quantitatively the impact of different data sources on the reconstruction task. Baseline models and the 4DVarNet framework are systematically compared. Second, we propose a quantitative analysis of the multi-modal approach. Different sources of information of the sea-surface state are characterized by heterogeneous spatio-temporal scales. These diverse data convey complementary information that could be fully embedded in the model with such a multi-modality aware scheme. Our framework can flexibly incorporate multiple sources of data and it proves superior to a modality-naive model. Third, we show that the presence of artificially injected bias in the low-resolution data to simulate random phase delay and re-modulation (often observed in real weather forecast products) may positively impact the generalization capabilities of the 4DVarNet framework. To complete the robustness analysis, we simulate the case in which in-situ observations cease to be available at test time, i.e. the model trained on high-resolution spatial and local observations is deployed to work on only spatial high-resolution observations. This experiment reveals that in-situ observation infrastructures installed in the close proximity of the coastline are more relevant for the sake of spatial reconstruction of wind speed. Finally, we perform a sensitivity analysis on the low and high-resolution data, changing the temporal sampling frequencies. 

The rest of this paper is structured as follows. Section~\ref{sec:problem-statement} states formally the problem and introduces the 4DVarNet scheme mathematical foundation. Section~\ref{sec:data} presents the dataset used and how the original data are prepared for our simulations. Section~\ref{sec:methods} details the application of the proposed method and all the experimental configurations. Section~\ref{sec:results} presents the results obtained and Section~\ref{sec:discussion} critically discusses these results.

\section{Problem statement}\label{sec:problem-statement}
The classical Variational Data Assimilation problem is based on a state-space formulation (see~\cite{evensen2009, carrassi2018data-assimilation}) that states both the system's state dynamics and the state observation process. In formulae
\begin{equation}\label{eq:state-space-formulation}
    \left\{
    \begin{array}{r c l}
         \displaystyle \dot{\mathbf{x}}(t) &=& \mathcal{M}(t, \mathbf{x}(t)) + \boldsymbol{\eta}(t) \\[2mm]
         \mathbf{y}(t) &=& \mathcal{H}(t, \mathbf{x}(t)) + \boldsymbol{\epsilon}(t) 
    \end{array}
    \right.
\end{equation}
where $\mathbf{x} \in X$ represents the state variable. In this case, $\mathbf{x}$ represents the complete high-resolution time series of wind speed. $\mathcal{M}$ is the operator that prescribes the system's dynamical behavior, $\mathbf{y} \in Y$ are the observations of the state variable, delivered by the observation model $\mathcal{H}$. This operator has partial access to the state variable which is the quantity that we want to obtain. The observation process delivers a low-resolution version of the state variable as well as masking an arbitrary number of its realizations. Both $\boldsymbol{\eta}$ and $\boldsymbol{\epsilon}$ are noise processes representing model and observations errors. The model~\eqref{eq:state-space-formulation} must be discretized. We define the flow operator associated to $\mathcal{M}$ as the one-step-ahead predictor of the state variable at time $t + \Delta t$ given its realization at time $t$,
\begin{equation}\label{eq:flow-operator}
    \Phi(\mathbf{x}(t + \Delta t)) = \mathbf{x}(t) + \int_{t}^{t + \Delta t} \mathcal{M}(t, \mathbf{x}(t)) \, dt
\end{equation}
Under the hypothesis of i.i.d. noise processes (\cite{carrassi2018data-assimilation}), the (weak-constraint) 4DVar problem of retrieving the complete time series of the state variable $\mathbf{x}$ is solved by inverting the forward processes stated in Equation~\eqref{eq:state-space-formulation}. This is accomplished by optimizing the following variational criterion
\begin{equation}\label{eq:var-cost}
    U_{\Phi}(\mathbf{x}, \mathbf{y}; \Omega) =
    \lambda_1 \, \sum_{t = 0}^{T} \| \mathcal{H}(t, \mathbf{x}(t)) - \mathbf{y}(t) \|^2 + \lambda_2 \, \sum_{t = 0}^{T} \| \mathbf{x}(t) - \Phi(\mathbf{x}(t)) \|^2  
\end{equation}
In this expression, $\| \cdot \|$ is the $L^2$ norm and $\Omega$ is the spatio-temporal domain defining the observation sampling pattern. The weights $\lambda_1$ and $\lambda_2$ are tunable parameters to weight each term. The variational cost is composed of a first data proximity term and a second regularization term, that enforces the compliance with the first equation in the model~\eqref{eq:state-space-formulation}. We obtain the optimal state variable as the minimizer of the variational cost
\begin{equation}\label{eq:var-cost-minimization}
    \hat{\mathbf{x}} = \arg \min_\mathbf{x} U_{\Phi}(\mathbf{x}, \mathbf{y}; \Omega)
\end{equation}
The spatio-temporal data sampling pattern is defined by the observation operator, which can act as a binary mask (\cite{fablet2021jointinterpolation}). The solution of the 4DVar problem applied to this case is the complete time series of high-resolution wind speed. The classical variational data assimilation schemes tackle this optimization problem with the adjoint method for the computation of the gradients of the cost function~\eqref{eq:var-cost} (\cite{Talagrand2015}). 

\section{Data}\label{sec:data}
We focus here on the exploitation of simulation datasets. We may emphasize that the direct exploitation of real observation datasets is not straightforward. It would imply a huge amount of work for mining and co-locating data on the same spatio-temporal grid, with no guarantee to build a sufficiently large and consistent dataset on a regional scale to run the targeted learning-based experiments. We then prioritize a preliminary study with simulations datasets. Recent studies (see~\cite{Febvre2023training}) also suggest the potential of such simulation-based datasets to train deep learning schemes which apply to real observation datasets.

\begin{figure}
    \centering
    \includegraphics[width=\textwidth]{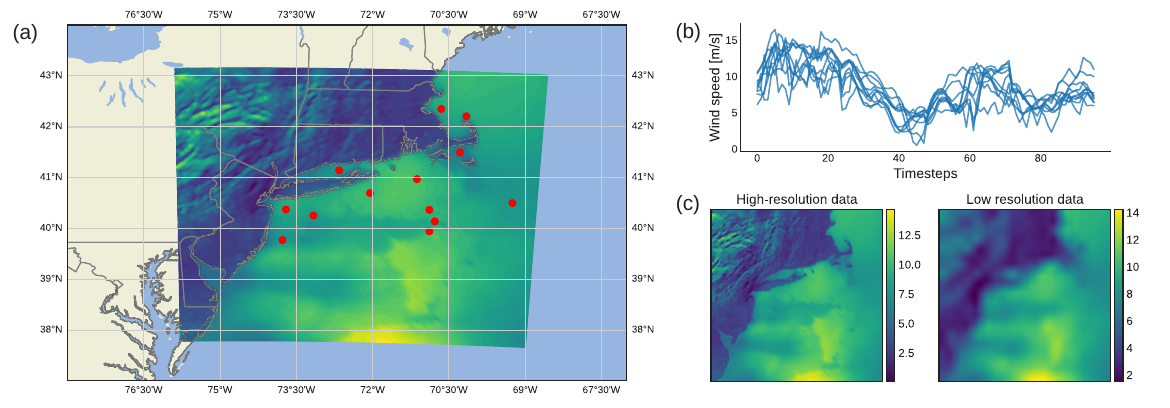}
    \caption{Dataset qualitative characteristics. Panel (a): Geographical region considered. The red markers represent the buoys positions. Panel (b): Sample in-situ time series pseudo-observations obtained from wind speed values at buoys positions. Panel (c): The downsampling-reinterpolation step to obtain low-resolution pseudo-observations from ground-truths. The spatial resolution for this case is 30 kilometers.}
    \label{fig:figure-dataset}
\end{figure}

We use the output of the RUWRF model (Rutgers University Weather Research and Forecast,~\cite{Optis2020validation}), based on the version 4.1.2 of the Weather Research and Forecast model (WRF, see~\cite{Powers2017weather}). The RUWRF is developed by the Rutgers University Center for Ocean Observing Leadership. This model runs a parent nest with resolution 9~\si{\kilo\meter} for a time interval of 120 hours and then a child nest with resolution 3~\si{\kilo\meter} out of 48 hours. The model is run daily. This implies a discontinuity in the data time series between 23:00 and 00:00 of two consecutive days. This motivates us to use 24 hours as reference time series length. The wind speed data are available in terms of the horizontal components. Our analyses target the modulus of wind speed. We process the components into the vector norm in order to treat the wind speed modulus. These wind speed fields are chosen to have a spatial extent of roughly 644 $\times$ 645~\si{\kilo\meter}, with a spatial resolution of $0.03^{\circ} \times 0.03^{\circ}$ (on average 3~\si{\kilo\meter}). The time window selected spans from 01/01/2019 at 00:00 to 01/01/2021 at 23:00. We prepare the wind speed fields and in-situ data as time series of length 24. Figure~\ref{fig:figure-dataset} displays examples of wind speed fields on the spatial region considered with associated pseudo-observations.

\subsection{Low-resolution data}\label{sec:data-lr}
We simulate low-resolution (LR) data similar to reanalysis data bases obtained using state-of-the-art data assimilation schemes (see~\cite{bannister2017, carrassi2018data-assimilation}). Such products typically resolve a wide range of spatial and temporal scales, depending on whether the numerical weather prediction (NWP) model is configured to address mesoscale, synoptic or climate scales phenomena. The spatial scales involves range between few tens to thousands kilometers. Likewise, temporal scales range from sub-daily to yearly. We adopt the following strategy. LR wind speed fields are obtained by down-sampling the original data. These fields are then re-interpolated on the reference grid to match the spatial resolution of the original data, as in Panel (c) Figure~\ref{fig:figure-dataset}. We choose to manufacture the LR data to have spatial resolution of 30 to 100~\si{\kilo\meter} and a time-step of 1 to 6 hours. This allows us to emulate products that resolve different spatio-temporal scales. The 30~\si{\kilo\meter} and 1~h configuration matches with the ECMWF ERA-5 data base (see~\cite{Hersbach2020-era5}). The configuration 100~\si{\kilo\meter} and 6~h matches with the previous ECMWF reanalyses data base ERA-interim (\cite{Dee2011}).

Reanalyses datasets may also involve local errors and biases imputable to Gaussian assumptions (\cite{carrassi2018data-assimilation}). These errors may relate to random delays in the weather forecast or random phase alteration. This kind of anomalies are observable in real-world NWP outputs, that may predict a given weather phenomenon at a given time, but this phenomenon is observed earlier or later, and with an intensity that may be different than the one predicted. In our experiments, we account for these possible biases by artificially injecting a phase delay or an amplitude re-modulation. Throughout the rest of this chapter, we refer to LR fields and NWP products (or pseudo-observations) interchangeably. 

\subsection{High-resolution data}\label{sec:data-hr}
We aim to mimick to some extent the sampling pattern of SAR satellite sensors (\cite{Monaldo2013seasat, Moreira2013tutorial}). We assume pseudo-SAR observations as noise-free HR snapshots of the sea surface wind speed. We focus here on the time sampling pattern. We assume that we are respectively provided by one and two HR observations over each 24-hour window. With a view to assessing the added value of a second SAR observation within a 24-hour window, we assume for the sake of simplicity that the two HR observations are sampled with a 12 hours delay.  

\begin{figure}[t]
    \centering
    \includegraphics[width=0.85\textwidth]{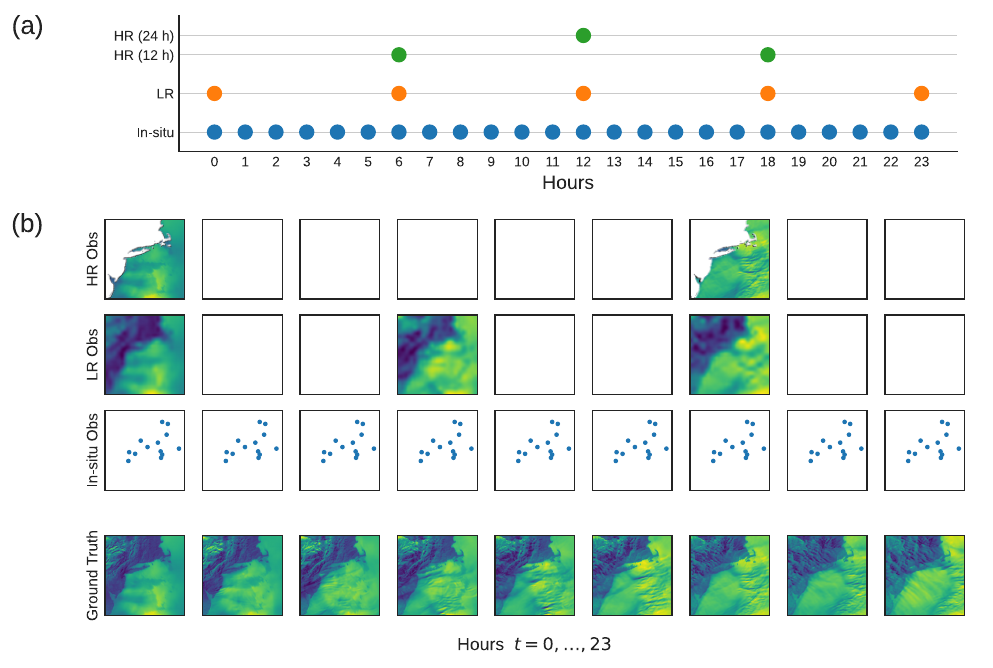}
    \caption{Panel (a): Temporal sampling patterns of the data used. The items ``HR (12~h)'' and ``HR (24~h)'' refer to the datasets in which the HR observations are simulated to have temporal frequency of 12 or 24 hours. These items refer to different experimental configurations and are depicted on the same plot for graphical convenience. Panel (b): An example of the dataset items. The temporal sampling frequencies of HR and LR fields are fictitious and aim to illustrate the dataset.}
    \label{fig:dataset-sfreq}
\end{figure}

Overall, we simulate here two HR observation datasets: for the first one, we provide a HR observation at the center of the 24-hour time window, namely at 12:00. For the second one, two HR observations are placed at 06:00 and 18:00, so to be 12-hours away. In the two cases the temporal sampling frequency of HR data is 24 and 12 hours respectively. In the experiments, we tested both these frequencies. Figure~\ref{fig:dataset-sfreq} gives a visual explanation of observations sampling patterns on a 24-hours time window. HR observations are masked in order to keep the wind speed information associated with the sea surface. This choice is motivated by the fact that SAR imagery can provide wind speed information by the sea surface roughness. This is not possible to achieve on the land surface. 

\subsection{In-situ time series}\label{sec:data-situ}
In-situ observation infrastructures (\cite{Gould2013situ}) provide real-time in-situ measurements. In-situ data have a fine temporal resolution and directly measure the quantity of interest, thus not being prone to any model error. The limitation of in-situ sensors is the deployment cost and limited spatial coverage. We simulate pseudo in-situ time series accounting for the positions of the weather buoys of the NOAA National Data Buoy Center network (website: \url{https://www.ndbc.noaa.gov/}), depicted in panel (a) of Figure~\ref{fig:figure-dataset}. Refer to~\cite{Green2006transitioning} for an overview of NOAA buoys installations. We select the buoys included in chosen region that were active in the chosen dataset temporal window. Using the positions of these buoys on the data grid, we keep the pixel values of the HR spatial fields and we extract these set of positions so to have multi-variate time series of point-wise wind speed values. As a result, we have thirteen buoys in the dataset. A portion of these time series is shown in panel (b) of Figure~\ref{fig:figure-dataset}.

\subsection{Preprocessing scheme}\label{sec:data-preprocessing}
The dataset is composed of $732$ time series of $24$ wind fields. We allocate the first $432$ series for the training set, the subsequent $200$ series for the test set and the last $100$ series for the validation set. In order to implement the simulations on LR data biased by random phase delays and amplitude remodulations we extract 36-hours time series in an early data processing stage. This choice is motivated by the fact that the hour 00:00 of each series may be modified by the random delay. If negative, this delay assigns to the field at 00:00 a value of the previous series. In this way, the early-stage series cover a time window from 18:00 of day $T-1$ to 06:00 of day $T+1$. This preprocessing step implies the loss of the first and last days of each data set. The training, validation and test set will then have lengths of $430$, $198$ and $98$, respectively. Once the LR bias has been injected, if the test case requires it, the 36-hours time series are cropped in order to have 24-hours long sequences starting at 00:00 and ending at 23:00 of day $T$. 

To sum up, the final dataset is composed of the three modalities mentioned above: the HR fields, the LR fields (with the injected bias if the training configuration requires) and the in-situ time series. From a practical point of view the dataset object provides a time series for each modality. Each time series is composed by 24 matrices having the shape of the spatial domain. The series are masked according to the temporal sampling frequency and the spatial features of each modality. For example, the in-situ observations are obtained by setting to zero the whole field except for the positions of the buoys. Panel (b) of Figure~\ref{fig:dataset-sfreq} gives a visual intuition of the data appearance. The three time series related to the pseudo-observations are obtained from the ground truths, the bottom series.

Data of the training, validation and test sets are normalized by a field-wise division by the standard deviation of the the respective set. For example, each field of the training set is divided by the standard deviation of the entire training set. Once the model is trained, data are de-normalized in order to evaluate the model performance in terms of physical dimensions.

\section{Methods}\label{sec:methods}
Here we detail the method outlined in Section~\ref{sec:problem-statement} and provide an explanation for the experimental configurations we test the model on.

\subsection{Trainable data assimilation scheme}\label{sec:tda-scheme}
The classical way to approach the problem stated in Section~\ref{sec:problem-statement} is to approximate the one-step-ahead predictor as defined in Equation~\eqref{eq:flow-operator} with Euler or Runge-Kutta schemes (\cite{runge1924, cash2003}). In the 4DVarNet scheme, we parameterize this operator with a trainable neural network. 
The observation operator $\mathcal{H}$ is a binary mask that enforces the spatio-temporal sampling pattern. For example, if the sampling frequency of HR fields is 12 hours, then the observation operator delivers the HR observations of the spatial wind speed at hours 06 and 18 for each daily time series. For HR spatial fields, the observation operator masks the land surface, in order to emulate more realistically the SAR products. For LR fields, the observation operator enforces and temporal sampling frequency and down-samples the original HR data. This reduces the problem to the task of learning an interpolation operator that fills the gaps imputable to the observation process and to improve the spatial resolution of the interpolated fields. In our case, the state variable $\mathbf{x}$ is the temporal sequence of HR wind speed field. The forward process $\mathcal{H}$ provides the observations $\mathbf{y}$ described in the previous section. More specifically, let $\mathbf{y}^{lr}$, $\mathbf{y}^{hr}$ and $\mathbf{y}^{situ}$ be the LR, HR and in-situ pseudo-observations respectively. Let $\Omega^{lr}$, $\Omega^{hr}$ and $\Omega^{situ}$ be the binary masks that identify the temporal sampling patterns associated to HR, LR and in-situ observations. Let $\Omega^{sl}$ be a mask that covers the pixel locations of the land surface. This mask enforces the SAR-like observations to be available only on the sea surface. Formally we can express the observation operator as follows
\begin{equation}\label{eq:mask-sampling-schemes}
    \mathcal{H}(\mathbf{x}) ~=~ 
    \left\{
        \begin{array}{l c l}
             \mathbb{I}^{hr} (\mathbf{x}) &=& \mathbf{y}^{hr}  \\[2mm]
             \mathcal{D} \circ \mathbb{I}^{lr} (\mathbf{x}) &=& \mathbf{y}^{lr} \\[2mm]
             \mathbb{I}^{situ} (\mathbf{x}) &=& \mathbf{y}^{situ}
        \end{array}
    \right.
\end{equation}
The symbols $\mathbb{I}^{hr}$, $\mathbb{I}^{lr}$ and $\mathbb{I}^{situ}$ represent the indicator functions associated with the spatio-temporal domains $\Omega^{hr} \cup \Omega^{sl}$, $\Omega^{lr}$ and $\Omega^{situ}$ respectively. The symbol $\mathcal{D}$ represents the downsampling-reinterpolation operation performed to obtain the LR fields. The objective is to retrieve the complete time series of HR surface wind speed using these partial observations. Recalling the form of the variational cost~\eqref{eq:var-cost}, restated in matrix form, this simple case of non-trainable data fidelty term (DFT) reduces to the optimization of the following cost function
\begin{equation}
    U_{\Phi}(\mathbf{x}, \mathbf{y}; \Omega) = \lambda_1 \, \| \mathbf{x} - \mathbf{y} \|^2_\Omega + \lambda_2 \, \| \mathbf{x} - \Phi(\mathbf{x}) \|^2
\end{equation}
We call this case ``single-modal DFT''. In a second case, we still have to enforce the sampling pattern with a binary mask but the occluded and down-sampled observations are treated with trainable operators. This ``multi-modal DFT'' learns feature maps from spatial and/or sequential observations and allows us to relate the state variable and the observations at a higher abstraction level that transcends the spatio-temporal features of the raw observations. We think that this is particularly relevant as spatial scales which characterize spatial fields and point-wise time series are drastically different. In this case the variational cost~\eqref{eq:var-cost} has an additional term, and can be stated as follows
\begin{equation}
    U_{\Phi}(\mathbf{x}, \mathbf{y}; \Omega) = \lambda_1 \, \| \mathbf{x} - \mathbf{y} \|_{\Omega}^2 + \lambda_1 \| f(\mathbf{x}) - g(\mathbf{y}) \|_{\Omega}^2 + \lambda_2 \| \mathbf{x} - \Phi(\mathbf{x}) \|^2
\end{equation}
In this equation, $f$ and $g$ are the neural networks. Depending on whether the observations $\mathbf{y}$ are spatial (HR wind fields) or sequential (point-wise time series), the networks classes $f$ and $g$ are 2D or 1D convolutional, respectively. The state variable $\mathbf{x}$ is intended to be a spatial field in any case. The observations $\mathbf{y}$ are multi-variate time series or spatial fields, or both, depending of the experimental configuration.

The trainable gradient solver $\Gamma$ is parameterized with a 2D convolutional LSTM (Long Short Term Memory, \cite{hochreiter1997}) network. The optimization of the state variable is achieved by an iterative application of the gradient solver to the gradients of the variational cost. The iterative rule to solve the optimization problem can be stated as follows
\begin{equation}
    \left\{
        \begin{array}{r c l}
             \mathbf{g}^{k}     &=& \nabla_\mathbf{x} \, U_{\Phi}(\mathbf{x}^{k}, \mathbf{y}; \Omega) \\[2mm]
             \mathbf{x}^{k + 1} &=& L \, \Gamma(\mathbf{g}^{k}, \mathbf{c}^{k}, \mathbf{h}^{k}) 
        \end{array}
    \right.
\end{equation}
where the $k$ superscript refers to the iteration, $\mathbf{c}$ and $\mathbf{h}$ are the cell and hidden states of the LSTM respectively and $L$ is a linear layer to map the output of the gradient solver to the dimensions of the state variable. 

The neural network parameterizations $\Phi$ and the gradient solver $\Gamma$ define the 4DVarNet end-to-end trainable architecture, denoted as $\Psi_{\Phi, \Gamma}$. The architecture can be seen as a learnable interpolation operator seeking for solutions that minimize the energy functional defined by the variational cost (\cite{fablet2021jointinterpolation}). 

\subsection{Direct learning-based inversion}\label{sec:dl-endtoend--approach}
Deep learning has recently been applied as a method to solve inverse problems (\cite{Ongie2020deep}) in the fields of medical imaging (\cite{Hyun2021deep}), computer tomography (\cite{Bubba2019learning}) computational photography (\cite{Chen2018learning}) and geosciences (\cite{Yu2021deep}). An alternative way to state the inversion problem stated in Section~\ref{sec:problem-statement} involves the direct application of deep learning models. This strategy allows to directly retrieve the state variable $\mathbf{x}$ as the output target of a multi-layered architecture as in the following expression
\begin{equation}\label{eq:direct-inversion}
    \mathbf{x} = f(\mathbf{y}; \bm{\theta})
\end{equation}
where $\bm{\theta}$ are the model $f$ parameters. The observations $\mathbf{y}$ follow the sampling patterns identified by the expression~\eqref{eq:mask-sampling-schemes}. The capability of deep networks to build effective feature maps and learn the input-output relationships from data allows to retrieve the system state directly, with no physical constraints. The difference between the inversion approach based on variational data assimilation methods is clear. A learning-based direct inversion as stated by a relationship as that stated in Equation~\eqref{eq:direct-inversion} does not require the knowledge of the dynamical model $\mathcal{M}$.

Deep learning methods can also be useful for super-resolution problems. This computer vision task aims to improve an image resolution. Previous work applied learning-based modelling to this end (\cite{Ducournau2016deep}). Impressive results are achieved by Generative Adversarial Networks (GANs,~\cite{Goodfellow2014generative}). Recent work applied GANs to the super-resolution of wind speed (\cite{stengel2020}), temperature (\cite{Lambhate2020super}) and precipitation~\cite{Leinonen2020stochastic}. However, GANs may be difficult to train due to the sensible choice of hyper-parameters and the model size. Recent advances in the deep learning field presented diffusion models (\cite{Croitoru2023diffusion}) as appealing alternatives for their efficiency and performance. Recent work applied diffusion models to high-resolution solar radiance forecast (\cite{Hatanaka2023diffusion}), seismic waves (\cite{Durall2023deep}) and medium-range weather forecast super-resolution (\cite{Chen2023swinrdm}). While generative deep learning-based super-resolution allows to generate graphically-realistic fields, our purpose is to apply deep learning modelling to reconstruct a sequence of wind speed fields as close as possible to reality. A wind speed field generated by a GAN or diffusion model may look more realistic than a direct inversion-produced one but there is no guarantee for the former to be physically plausible.

\subsection{Learning scheme}\label{sec:learning-scheme}
The training loss chosen to optimize the model parameters is a mean squared error (MSE) composed of terms related to LR and HR reconstructions. The input data to the model are the LR and the anomalies of the spatial fields. We define the anomaly as the difference between the HR and the LR fields. In formulae,
\begin{equation}
    \mathbf{y}^{an} = \mathbf{y}^{hr} - \mathbf{y}^{lr}
\end{equation}
The state variable $\mathbf{x}$ and the observations $\mathbf{y}$ are composed of both the LR and anomaly fields. The state variable and observations tensors are processed in order to host the LR information and two times the anomaly information. The first instance of the anomaly is used in the 4DVarNet computations and the second is used as output. This choice is motivated by the fact that the part processed by the 4DVarNet operators may have undesired artifacts in the reconstruction (\cite{Beauchamp20224dvarnet}). The training loss comprehends also one term which enforces the spatial gradients of the HR reconstructions and ground-truth data to be similar. Let $\mathbf{u}$ represent the ground-truths used in the problem design and $\hat{\mathbf{x}}$ the model output. The training loss can be formally stated as 
\begin{equation}\label{eq:training-loss}
    \mathcal{L}(\mathbf{u}^{hr}, \hat{\mathbf{x}}^{hr}) = \frac{1}{M}\sum_{i = 0}^{M} \sum_{t = 0}^{T} \left\{ \, \| \mathbf{u}_{it}^{lr} - \hat{\mathbf{x}}_{it}^{lr} \|^2 + \| \mathbf{u}_{it}^{hr} - \hat{\mathbf{x}}_{it}^{hr} \|^2 + \| \nabla \, \mathbf{u}_{it}^{hr} - \nabla \, \hat{\mathbf{x}}_{it}^{hr} \|^2 \, \right\} 
\end{equation}
The symbol $\nabla$ identifies the spatial gradients of a field w.r.t. the spatial coordinates. The terms of the loss function involving gradients are needed to enforce the spatial variation patterns of the reconstructions to match those of the ground-truths. The training criterion~\eqref{eq:training-loss} is complemented by a $L_2$ regularization term to prevent overfitting.

\subsection{Numerical implementation}\label{sec:numerical-implementation}
The optimization of the cost function~\eqref{eq:training-loss} is done with the Adam algorithm (\cite{adam14}). The model is trained for 50 epochs. The weights configuration used for the test stage is selected according to the validation loss during training. Table~\ref{tab:hyperparams} resumes the learning rates and $L_2$ regularization weight decay coefficients for the trainable models deployed. Our choice is to set the weights of the variational cost terms $\lambda_{1,2}$ to trainable parameters. In the following we report briefly the neural architectures used for this work. The interested reader may refer to the code available at \url{https://github.com/CIA-Oceanix/4DVN-MM-W2D/tree/main} for the complete and detailed models design. We provide the motivation for the design choice of the model $\Phi$ in Appendix~\ref{app:app-model-phi}.
\begin{itemize}
    \item[{\textbullet}] The operator $\Phi$ is parameterized by a 2D convolutional network with two layers. The kernel size is chosen to be $5$ with padding $2$ (both are isotropic). The input tensor is a concatenation of three times the spatial fields (LR and anomalies), so the temporal dimension becomes $72$. For this reason, the input channels are set to $72$. The first layer squeezes the channels to $32$ and the second layer expands the channels to $72$. The model does not have non-linearities. 
    \item[{\textbullet}] The operator $\Gamma$ is parameterized by a 2D convolutional LSTM. The dimension of the hidden cell is set to $100$. The gates layer is parameterized by a 2D convolutional layer with kernel size $3$ and padding $1$. The input channels are $96$ and the gates layer expands this number to $384$. A linear 2D convolutional layer (kernel size $1$ and no padding) reshapes the LSTM output to the original input size.
    \item[{\textbullet}] The models $f$ and $g$ are involved in the multi-modal DFT case. They process the system state and observations respectively. The model for the system state $f$ is a 2D convolutional model. According to whether the observations are spatial wind fields or in-situ time series, the model $g$ is a cascade of 2D convolutional layers and Average pooling layers or a 2D convolutional layer. In the case where both spatial and time series are involved, two different models $g$ are deployed. The extracted features maps are flattened to be compared by the variational cost observations term.
\end{itemize}

\begin{table}[t]
    \centering
    \begin{tabular}{c c c c c}
        \toprule[1pt]
                           & Operator $\Phi$            & Solver $\Gamma$          & Models $f$, $g$  & Weights $\lambda_{1,2}$ \\
        \midrule
        Learning rate      & $5 \cdot 10^{-5}$ & $9 \cdot 10^{-5}$ & $10^{-4}$ & $10^{-4}$ \\
        $L_2$ weight decay & $10^{-7}$         & $10^{-8}$         & $10^{-7}$ & $10^{-5}$ \\
        \bottomrule[1pt]
    \end{tabular}
    \caption{Hyper-parameters used for the simulations. Depending on the case, some columns may not apply. For example, in the case of non-trainable DFT, the column $f$, $g$ is not considered.}
    \label{tab:hyperparams}
\end{table}

The simulation are run on a machine mounting 8 Nvidia A100-SWM4 graphical processor units. These units have 80~GB memory and GA100 graphics processors with 1593~MHz memory clock speed. Training times are reported in Appendix~\ref{app:app-model-phi}.

\section{Results}\label{sec:results}
Results articulate as follows. We detail the baseline models that we choose to compare the 4DVarNet framework with. We provide the technical details about the evaluation metrics and we detail each experimental configuration. The baseline models and the 4DVarNet are compared. The analyses on robustness against biased LR fields, buoys sensitivity and scale analyses are made using the best model as of the benchmark configuration. The results of the analyses on high and low-resolutions sensitivity are obtained by models trained on the specified data sampling configurations.

\subsection{Evaluation framework}\label{sec:eval-framework}
The reconstruction performance is evaluated quantitatively in terms of root mean squared error (RMSE) between ground-truth data and the output of each model. The reconstruction error is evaluated on the complete region considered and on both land and sea areas separately. We perform 10 runs for each configuration and we compare the median reconstruction against the ground-truth. This procedure can be viewed as a \emph{voting} average to aggregate an ensemble of models (\cite{Dietterich2000ensemble}). Ensemble Machine Learning methods construct a group of independently trained models. The overall ensemble output has an overall reduced variance (\cite{Rincy2020ensemble}). The aggregated median reconstruction of the models ensemble is obtained as
\begin{equation}
    \hat{\mathbf{x}}_{\text{median}} = \text{Median} \, (\; \{ \hat{\mathbf{x}}_n \,; \; 0 \le n < N_{\text{Runs}} \} \;)
\end{equation}
where $\hat{\mathbf{x}}_n$ is the run-wise model output. The model performance is evaluated against a chosen baseline. The next Section provides a detailed overview of models and baselines. We may define a relative gain to compare the percentage improvement of each model w.r.t. the selected baseline. This gain in defined as 
\begin{equation}\label{eq:relative-gain}
    \eta = \left( 1 - \frac{p_M}{p_B} \right) \times 100
\end{equation}
where $p_M$ and $p_B$ are the reconstruction RMSEs of a given model $M$ and the reference baseline $B$ respectively. The performance $p_M$ is referred to the ensemble reconstruction. This gain may attain virtually any real value. If it is negative, we refer to it as ``degradation'' as the performance of a model $M$ may be worse than the baseline $B$ performance.

\subsection{Benchmark analysis results}\label{sec:exp-config}
In order to benchmark the 4DVarNet model we propose the following configuration of available data: (i) LR wind speed fields simulating a NWP product with a sampling frequency of 6 hours; (ii) HR pseudo-observations that simulate the satellite images, available with a temporal sampling frequency of 12 and 24 hours, respectively at hours 06:00 and 18:00 for the first case and 12:00 for the second; (iii) in-situ time series, with hourly resolution. We identify four test cases given by the combination of these data modalities. Table~\ref{tab:benchmark-exp-config} resumes these configurations. The 4DVarNet scheme is compared with two baseline models. The first baseline used is a temporal interpolation of LR fields, with no HR observations. In this case, the interpolation operator is not trainable. The second baseline model is a learning-based direct inversion (cfr. Section~\ref{sec:dl-endtoend--approach}) applied to the data configurations of Table~\ref{tab:benchmark-exp-config}. The target state variable $\mathbf{x}$ is obtained by the direct application of a trainable operator to the observed data as stated by Equation~\eqref{eq:direct-inversion}. The direct inversion for the data configuration SR solves a \emph{super-resolution} task. This means that the model is trained to retrieve the finer-scale information using only the LR fields. Let $B_0$ and $B_1$ represent the interpolation and direct inversion baselines respectively. Let $M_s$ and $M_m$ denote the 4DVarNet framework in the single-modal and multi-modal DFT settings respectively (cfr. Section~\ref{sec:tda-scheme}). In the following, the notation $\{B_0, B_1, M_s, M_m\}$-$\{\text{SR}, \text{C1}, \text{C2}, \text{C3}\}$ is used to identify the combination of model and experimental configuration. This convention is used hereafter in the rest of the paper. The 4DVarNet dynamical operator $\Phi$ shares the same architecture of the trainable direct inversion model.

\begin{table}[t]
    \centering
    \begin{tabular}{c c c c}
    \toprule[1pt]
         \textbf{Configuration} & LR data & HR data & In-situ data \\
         \midrule
         \textbf{SR} & {\color{dgreen}{6~h}} & \color{dred}{\xmark} & \color{dred}{\xmark} \\ 
         \textbf{C1} & {\color{dgreen}{6~h}} & {\color{dgreen}{\{12, 24\}~h}} & \color{dred}{\xmark} \\
         \textbf{C2} & {\color{dgreen}{6~h}} & \color{dred}{\xmark} & {\color{dgreen}{1~h}} \\ 
         \textbf{C3} & {\color{dgreen}{6~h}} & {\color{dgreen}{\{12, 24\}~h}} & {\color{dgreen}{1~h}} \\
    \bottomrule[1pt]
    \end{tabular}
    \caption{Experimental data configurations. The curly brackets for the cases C1 and C3 represents the set to HR sampling frequencies inspected.}
    \label{tab:benchmark-exp-config}
\end{table}

\begin{figure}[t]
    \centering
    \includegraphics[width=\textwidth]{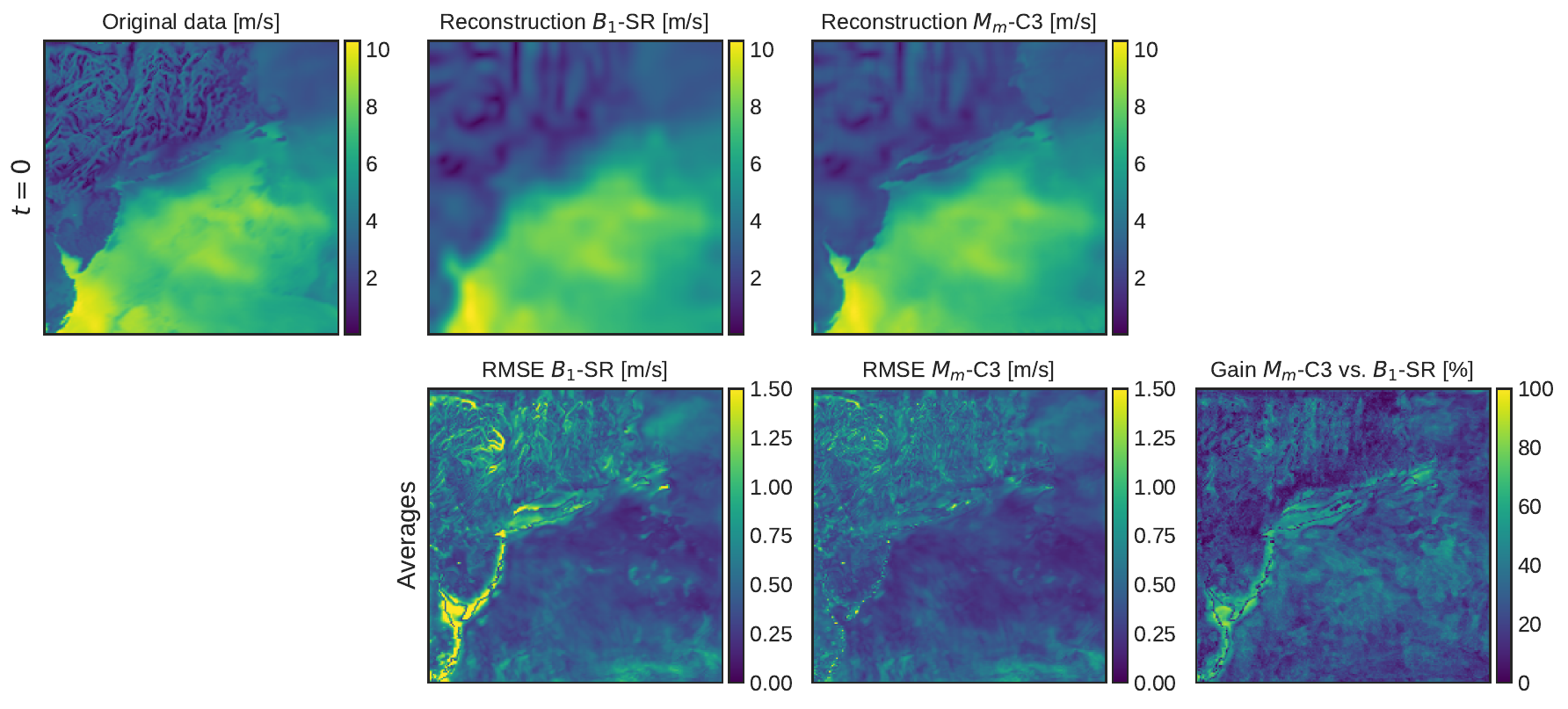}
    \caption{First row: (left) Original data, (middle) reconstruction of the $B_1$-SR baseline, (right) reconstruction of the $M_m$-C3 4DVarNet. Second row: (left) Map of average MSE related to the $B_1$-SR baseline, (middle) $M_m$-C3 4DVarNet and (right) map of the average relative gain of the 4DVarNet w.r.t. the baseline. The temporal sampling frequency for high-resolution fields is 12 hours. The two rows are displaced in order for the baseline and model reconstructions and error maps to match vertically.}
    \label{fig:fig-reconstructions-errors-gain}
\end{figure}

Table~\ref{tab:results} reports systematically the simulation results considering the data configurations mentioned above. The model $M_m$-C3 4DVarNet is the configuration that gives the best reconstruction performance. In Figure~\ref{fig:fig-reconstructions-errors-gain} the first row shows original data and reconstructions examples of both the baseline $B_1$-SR and the model $M_m$-C3. The second row shows the average RMSE map of $B_1$-SR and $M_m$-C3 (first and second bottom panels) and the average relative gain of $M_m$-C3 w.r.t. $B_1$-SR. This result may not raise any surprise, since the more the data used the more information is available for the reconstruction. Nevertheless, a very interesting result concerns the performances of the two instances of the 4DVarNet scheme, $M_s$ and $M_m$. Despite the model $M_s$ outperforms the direct inversion baseline, the addition of in-situ observations in the configuration C3 does not seem to be beneficial w.r.t. the configuration C1. Contrarily, the $M_m$ model benefits more from in-situ time series in configuration C3. The reconstruction performance is roughly $2~\%$ superior w.r.t. the case C1 for both the temporal frequencies. This result proves the capability of an explicit multi-modality processing to make the most out of both the sources of HR information. In addition, we compare the gain of the configurations $M_m$-C1 and $M_m$-C3 with the HR sampling frequencies of 12 and 24 hours respectively. This comparison aims to answer the question of whether it is preferable one supplementary HR snapshot or the in-situ time series. The average difference between the two performance levels is $1.13~\%$ in favor of the configuration $M_m$-C3 with HR sampling frequency of 24 hours. The same comparison for the $M_s$ model gives an average difference of $-0.12~\%$. This results show that the \emph{efficiency} of the multi-modal $M_s$ model allows to better use information coming from temporally rich in-situ time series. 

\begin{table}[t]
    \centering
    \begin{tabular}{c c c c c c c c c}
         \toprule[1pt]
         \multicolumn{2}{c}{Model} & $\omega_t^{hr}$ & \multicolumn{2}{c}{Full} & \multicolumn{2}{c}{Sea} & \multicolumn{2}{c}{Land} \\
         \phantom{textfillh} & & \phantom{textfillh} & RMSE & Gain & RMSE & Gain & RMSE & Gain \\
         \midrule
         $B_0$ &                  &  & $1.1234$ & & $1.1384$ & & $1.0983$ & \\
         \midrule
         \multirow{6}{*}{$B_1$}     & SR & --   & $0.9960$ &        & $0.9817$ & & $1.0192$ & \\
                                    & C1 & 12~h & $0.9605$ & $3.56$ & $0.9389$ & $4.36$ & $0.9951$ & $2.36$ \\
                                    & C1 & 24~h & $0.9741$ & $2.20$ & $0.9555$ & $2.67$ & $1.0040$ & $1.49$ \\
                                    & C2 & --   & $0.9957$ & $0.03$ & $0.9814$ & $0.03$ & $1.0187$ & $0.05$ \\
                                    & C3 & 12~h & $0.9571$ & $3.91$ & $0.9341$ & $4.85$ & $0.9938$ & $2.49$ \\
                                    & C3 & 24~h & $0.9711$ & $2.50$ & $0.9515$ & $3.08$ & $1.0024$ & $1.65$ \\
        \midrule
        \multirow{5}{*}{$M_s$}      & C1 & 12~h & $0.9000$ & $9.64$ & $0.8930$ & $9.04$ & $0.9114$ & $10.58$ \\
                                    & C1 & 24~h & $0.9012$ & $9.52$ & $0.8953$ & $8.80$ & $0.9108$ & $10.64$ \\
                                    & C2 & --   & $0.9619$ & $3.42$ & $0.9508$ & $3.15$ & $0.9798$ & $3.870$ \\
                                    & C3 & 12~h & $0.8999$ & $9.65$ & $0.8939$ & $8.94$ & $0.9096$ & $10.75$ \\
                                    & C3 & 24~h & $0.9015$ & $9.49$ & $0.8958$ & $8.75$ & $0.9107$ & $10.65$ \\
        \midrule
        \multirow{5}{*}{$M_m$}      & C1 & 12~h & $0.8802$ & $11.63$ & $0.8695$ & $11.43$ & $0.8974$ & $11.95$ \\
                                    & C1 & 24~h & $0.8907$ & $10.57$ & $0.8836$ & $9.99$ & $0.9022$ & $11.48$ \\
                                    & C2 & --   & $0.9197$ & $7.66$  & $0.9207$ & $6.21$ & $0.9180$ & $9.93$ \\
                                    & C3 & 12~h & $0.8617$ & $\mathbf{13.48}$ & $0.8481$ & $\mathbf{13.61}$ & $0.8836$ & $13.30$ \\
                                    & C3 & 24~h & $0.8692$ & $12.73$ & $0.8606$ & $12.34$ & $0.8832$ & $\mathbf{13.34}$ \\
        \bottomrule[1pt]
    \end{tabular}
    \vspace{0.25cm}
    \caption{Benchmark test results. $B_0$: interpolation. $B_1$ learning-based direct inversion. $M_s$ and $M_m$: 4DVarNet model with non-trainable and trainable observation operator, respectively. The gains are referred to the $B_1$-SR configuration and are computed as in Equation~\eqref{eq:relative-gain}. RMSE is expressed in \si{\meter\per\second} and relative gain in percentage. The six three columns report the RMSE and relative gain for the full region, the sea and land portions respectively. The symbol $\omega_t^{hr}$ refers to the temporal sampling frequency of high-resolution fields.}
    \label{tab:results}
\end{table}

\begin{figure}
    \centering
    \includegraphics[width=0.75\textwidth]{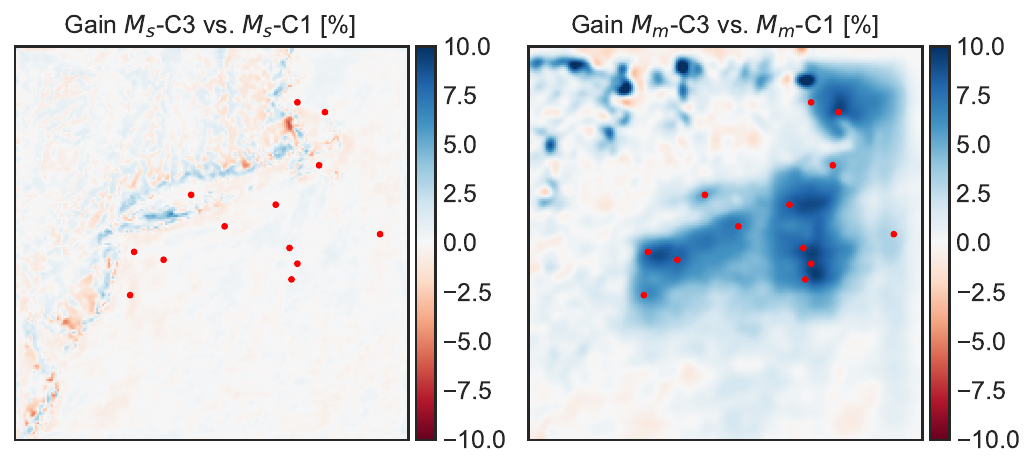}
    \caption{Maps of average relative gains. Left panel: Non-trainable observation operator. Average gain of model $M_s$-C3 w.r.t. $M_s$-C1. Right panel: Trainable observation operator. Average gain of model $M_m$-C3 w.r.t. $M_m$-C1. The temporal sampling frequency for high-resolution fields is 12 hours.}
    \label{fig:avg-gain-maps}
\end{figure}

In Figure~\ref{fig:avg-gain-maps} we report the visual representation of the improvement due to the multi-modality of the DFT by comparing the reconstruction gains of cases C3 and C1. For this visualization, the temporal sampling frequency chosen is 12 hours. The left panel represents the average gain of the model $M_s$-C3 w.r.t. $M_s$-C1, and the right panel is the average gain of model $M_m$-C3 w.r.t. $M_m$-C1. Intriguingly, the gains are not only restricted to areas surrounding the buoys but extend to regions of about one order of magnitude larger than the spatial scale of a local point-wise measurement. 
This improved reconstruction can even reconstruct the profile of the coastline. This result shows that a trainable multi-modal approach may incorporate the heterogeneous information of in-situ observations and spatial HR observations. The local small-scale information is learned and reused 
to a larger scale, which is defined by the spatial HR data.

\subsection{Biased low-resolution data}\label{sec:lr-bias}
The prediction made by a NWP system and the true realization of the phenomenon may differ in both timing and intensity. 
To simulate this scenario, the LR fields are biased by either a random phase delay $\Delta t$ in the interval $[-4, +4]$ hours or a random amplitude re-modulation $\alpha$ in the interval $[0.5, 1.5]$. This LR data modification is performed randomly at train time. More specifically, each field $\mathbf{y}^{lr}(t)$\footnote{The time step $t$ complies with the temporal sampling frequency prescribed by the observation process~\eqref{eq:mask-sampling-schemes}.} is modified by a randomly chosen phase delay or remodulation. Formally
\begin{equation}
    \mathbf{y}^{lr}(t) = 
    \left\{
    \begin{array}{r l}
         \mathbf{y}^{lr}(t + \Delta t) & \text{with } \Delta t \sim \mathcal{U}(-4, +4) \\[2mm]
         \alpha \, \mathbf{y}^{lr}(t)  & \text{with } \alpha \sim \mathcal{U}(0.5, 1.5) 
    \end{array}
    \right.
\end{equation}
The symbol $\mathcal{U}$ represents the uniform probability distribution. Interestingly, this procedure has a clear resemblance with dynamic data augmentation (\cite{Xu2021classification}). The model $M_m$-C3 with HR sampling frequency of 12 hours is trained with these two LR data modification separately. HR spatial and in-situ observations are not modified. We expect a degradation in the reconstruction performance, but the foremost interest of this analysis is to prove the robustness of the trainable variational scheme against the presence of model errors in the LR pseudo-observations. 
\begin{figure}[t]
    \centering
    \includegraphics[width=0.85\textwidth]{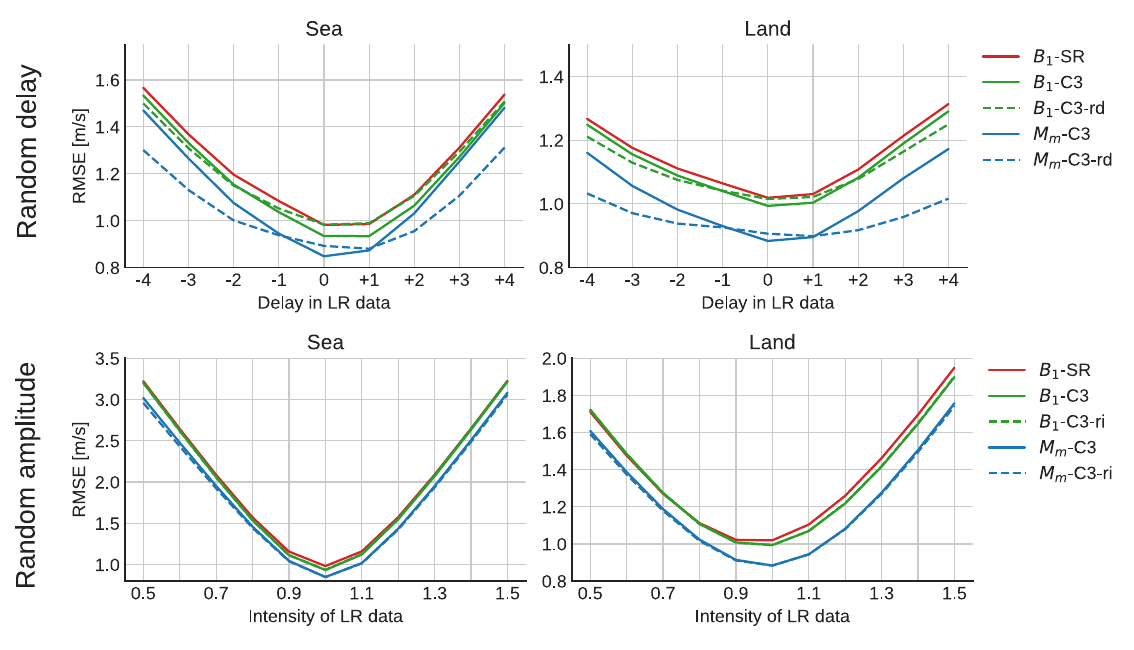}
    \caption{Top row: test case of LR simulated delay. Bottom row: test case of LR simulated re-modulation. 
    The suffixes ``-rd'' and ``-ri'' identify the models trained in case of random delay and re-modulation, respectively. The experimental configuration used for this experiment is the case C3 for the $M_m$ model and the SR, C3 cases for the $B_1$ baseline. The temporal sampling frequency of high-resolution fields is 12 hours.}
    \label{fig:lr-bias}
\end{figure}
The test procedure is performed as follows. The trained model is evaluated on an ensemble of modified test sets, where each of these sets is obtained modifying all the LR elements by one constant biasing value. For example, in the case of random delay the first test set has its LR elements modified by the delay $-4$ hours. The second test set's LR elements are modified by $-3$ hours and so on. The same method is applied for the case of the remodulation bias. This procedure gives an ensemble of 9 and 11 performance metrics for the delay and remodulation cases.

Figure~\ref{fig:lr-bias} reports the curves of reconstruction error as functions of the bias magnitude for both the delays (first row) and re-modulations (second row). Intriguingly, the model $M_m$-C3 trained on biased LR data outperforms the model trained on unbiased LR data for large biases values. This result can be interpreted as the model capability to learn the LR data correction, thanks to the trainable multi-modal observation operator that extracts meaningful features from the HR observations. This is particularly clear in the case of random delay, while for random re-modulation the effect is milder. This may be explained noting that model intensity errors in the range $\left[ 0.5, 1.5 \right]$ are not strong enough to spoil the results. Moreover, such biases may, contrarily, have a positive regularizing effect on the model training.

\begin{figure}[t]
    \centering
    \includegraphics[width=0.65\textwidth]{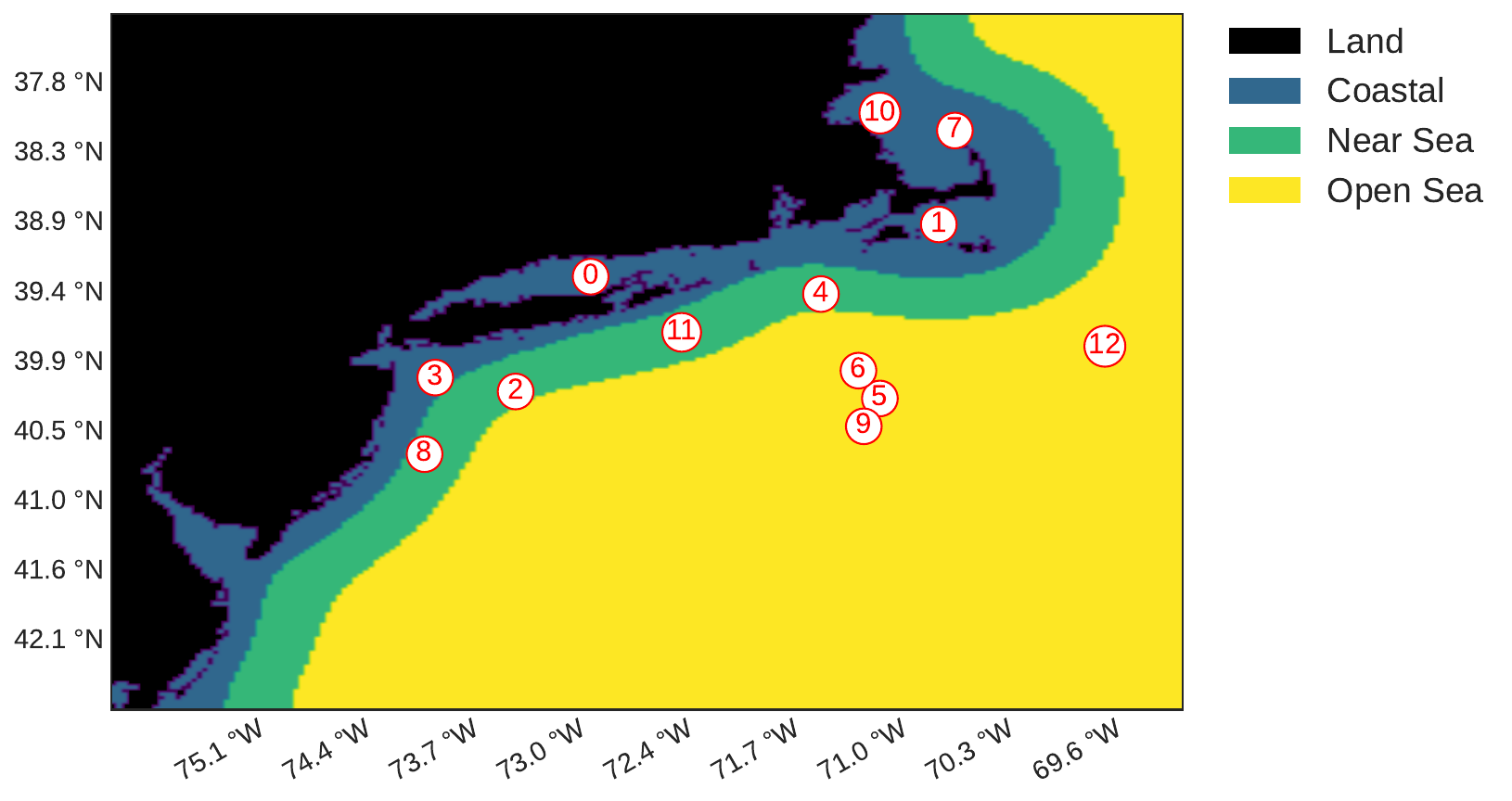}
    \caption{Division of the region of interest in the three zones that host the three buoys groups. The circled numbers represent the buoys identifiers on the respective buoy position.} 
    \label{fig:buoys-zones}
\end{figure}

\subsection{Buoys sensitivity analysis}\label{sec:buoys-sensitivity}
In this work we focus on the added-value of having in-situ sensors that measure directly the variable of interest. In real-world applications it is not possible to deploy a very dense network of observation infrastructures as such installation operations are expensive and demanding. In the following experiment, we aim to evaluate the impact of missing buoys on test stage using a model trained with both HR spatial wind fields and in-situ measurements. We use the best model of the case described in Section~\ref{sec:exp-config}, considering both the sampling frequencies of 12 and 24 hours for the HR spatial fields. The choice of performing the test with both the sampling frequencies is motivated by the necessity to assess the impact of in-situ observations related to two levels of HR spatial information availability. We set up two test cases. In the first case, we test the model removing one single buoy to see whether some buoys are more relevant for the reconstruction task. This test is repeated for each of the 13 buoys of the observation network. In the second case, we identify three zones on the region on interest, as illustrated in Figure~\ref{fig:buoys-zones}. A first zone is the strict proximity of the coastline, the second is an intermediate buffer zone between the coast and the open sea and the third zone is the open sea, from approximately 76 kilometers from the coastline. The objective of this second test case is to check whether one group of buoys provides more meaningful measurements for the task of wind speed reconstruction on a mixed land-sea region.

\begin{figure}[t]
    \centering
    \includegraphics[width=0.85\textwidth]{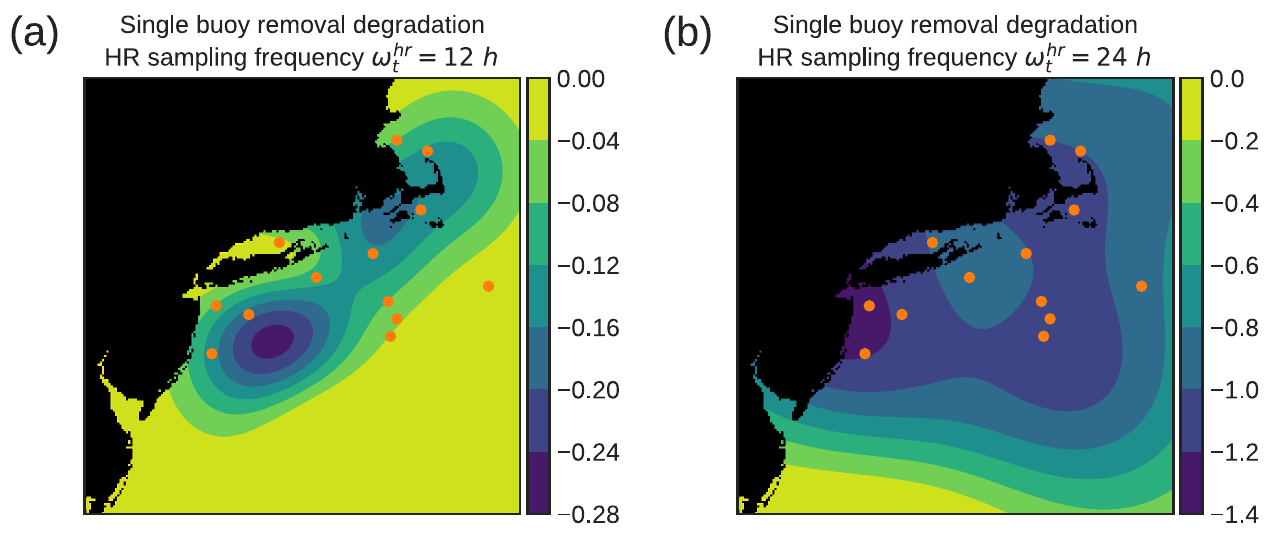}
    \caption{Single buoys removal tests. Degradation maps obtained interpolating spatially the buoy-wise degradation values. The interpolation is made with a Gaussian process regression. Panel (a): HR temporal sampling frequency of 12 hours. Panel (b): HR temporal sampling frequency 24 hours.}
    \label{fig:buoys-sensitivity-singlebuoy}
\end{figure}

\begin{table}
    \centering
    \begin{tabular}{c c c}
         \toprule[1pt]
         \phantom{texttofillwidth} & \multicolumn{2}{c}{Average degradation}     \\
         \textbf{Zone} & $\omega_t^{hr} = 12$~h & $\omega_t^{hr} = 24$~h \\
         \midrule
         Coastal       & $-0.69~\%$             & $-1.61~\%$             \\
         Near-sea      & $-0.66~\%$             & $-1.48~\%$             \\
         Open Sea      & $-0.52~\%$             & $-1.38~\%$             \\
         \bottomrule[1pt]
    \end{tabular}
    \caption{Average degradation values of the case of buoys group removal.}
    \label{tab:buoys-degradation-buoysgroups}
\end{table}

We start the analysis with the results related to the case of single buoy removal, cfr. Section~\ref{sec:buoys-sensitivity}. In order to visualize the degradation trend as a function of the single buoys, we produce a fictitious map as follows. The starting point is a blank matrix having the same dimensions as the input spatial data. We can identify on this map the buoys locations, as shown in Figure~\ref{fig:buoys-zones}. On the point associated with the buoy $n$, with $n = 1, \dots, 13$, we set the value of degradation due to the removal of the buoy $n$. In this way we obtain a group of discrete values of degradation. These values are interpolated spatially with a Gaussian process (\cite{Liu2018gaussian}). The resulting map can be interpreted as the individual buoy responsibility to degradation. Panel~(a) of Figure~\ref{fig:buoys-sensitivity-singlebuoy} displays the degradation map for the case of HR spatial fields temporal frequency of 12 hours, i.e. a HR field at hours 06 and 18. The degradation is mainly concentrated on a spatial belt encompassing the intermediate and coastal regions. This means that the most relevant buoys for the reconstruction task are those located in the strict proximity of the coast. The degradation map changes in the case of HR fields temporal frequency of 24 hours, see Panel~(b) of Figure~\ref{fig:buoys-sensitivity-singlebuoy}. In-situ observations are more that $1~\%$ more relevant as the HR observation regime is halved. In other words, the fine-scaled information provided by in-situ data is exploited more extensively by the model. In addition, the importance of the in-situ data is more distributed among all the buoys and there are not particularly preferred regions. We discuss now the results of the buoys groups removal case. Again, the average degradation values are larger in the case of HR sampling frequency of 24 hours. The reason for this is again the relevance of HR information of in-situ data. The difference w.r.t. the single buoy removal tests is the constant tendency of degradation, which attains larger values for the coastal region buoys and progressively decreases in the open sea direction. Table~\ref{tab:buoys-degradation-buoysgroups} reports values of error metrics, which may be explained as follows. The coastal buoys are placed on the area that experiences the most \emph{shelter effects}, involving strong wind speed gradients associated with the presence of the morphology of the coast (\cite{Cathelain2023high, Schulz2022coastal}). These results mean that the coastline proximity is the most sensitive region for the wind speed reconstruction task. The largest impact of in-situ observations is associated to this particular zone.

\subsection{Spatial fields resolution sensitivity analysis}\label{sec:spatial-fields-resolution-sensitivity-analysis}
To conclude this work, we propose an aggregated analysis on the impact of each data source on the reconstruction performance of the $M_m$ model. The tests involve the comparison between the C1 and C3 configurations (in-situ time series), the LR data divided in four groups as in Table~\ref{tab:hr-lr-st-res}. The group A corresponds to the LR configuration used in the test cases detailed above. Using the best benchmark model, for each group we test both the C1 and C3 configurations as described above and both the 12 hours and 24 hours temporal resolutions for HR spatial fields. These additional tests allow to better appreciate the impact of the temporal and spatial resolution of spatial fields, both HR and LR. 

\begin{table}[h]
    \centering
    \begin{tabular}{c c c}
    \toprule[1pt]
                  & Spatial resolution & Temporal resolution \\
         \midrule
         \textbf{Group A} & 30~km  & 6~h \\ 
         \textbf{Group B} & 30~km  & 1~h \\
         \textbf{Group C} & 100~km & 6~h \\ 
         \textbf{Group D} & 100~km & 1~h \\ 
    \bottomrule[1pt]
    \end{tabular}
    \caption{Combinations of the HR and LR fields spatio-temporal resolutions.}
    \label{tab:hr-lr-st-res}
\end{table}

\begin{figure}[t]
    \centering
    \includegraphics[width=\textwidth]{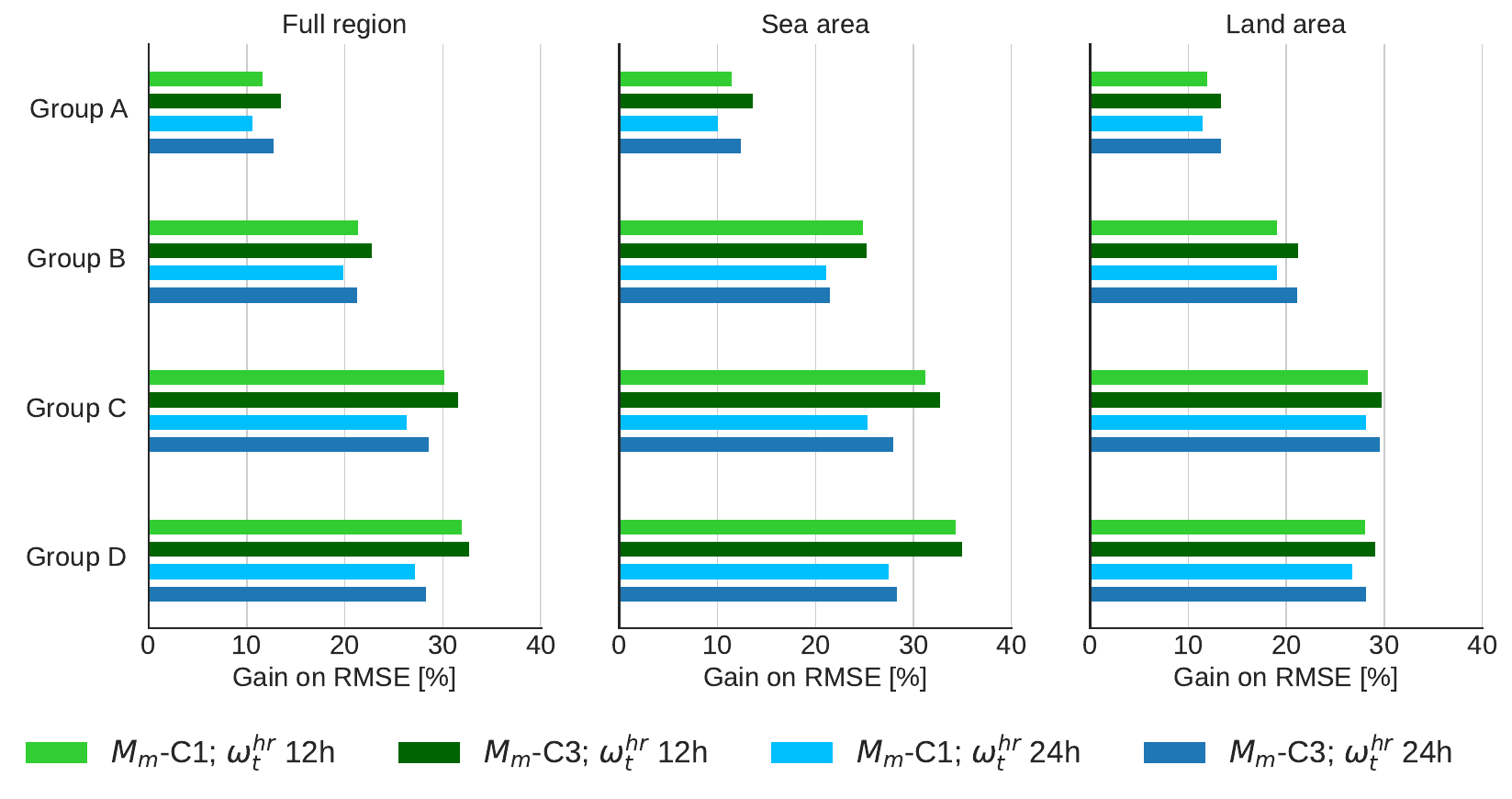}
    \caption{Sensitivity analysis results. The labels on the left identify the groups of LR spatio-temporal resolutions. The performance metrics illustrated are relative gains on RMSE referred to the $B_1$-SR baseline applied to each LR combination. The symbol $\omega_t^{hr}$ refers to the high-resolution spatial fields sampling frequency.}
    \label{fig:lr-hr-sensitivity-analysis}
\end{figure}

Figure~\ref{fig:lr-hr-sensitivity-analysis} visualizes the results of the performance of $M_m$ in all the configurations mentioned above. There are two evident remarks. The first one is that the difference between C3 and C1 is larger for the configurations where LR fields have temporal resolution of 6 hours. This can be explained as follows: if LR data have temporal resolution of 1 hour, the dynamical information of in-situ time series is partially hidden by the larger temporal availability of LR fields. Again, this result supports the importance of in-situ observations as source of time-dependent information. The second remark concerns the spatial resolution of LR data. For the higher resolution of 100~km, the fine-scale information delivered by both the spatial fields and the in-situ data are more largely exploited by the multi-modal trainable observation operator of the 4DVarNet. 

\begin{figure}
    \centering
    \includegraphics[width=0.75\textwidth]{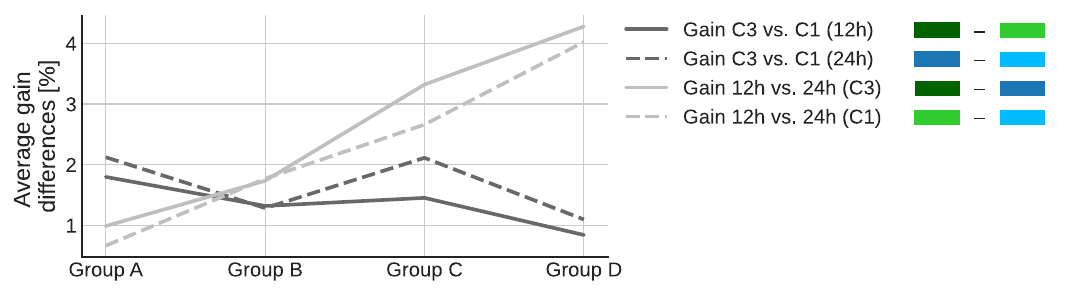}
    \caption{Analysis on the gains referred to the contributions of either the in-situ time series and a finer temporal resolution of high-resolution wind speed fields. The gains are referred to the $B1$-SR baseline associated to the low-resolution spatio-temporal resolutions of each group. The groups name convention follows the definitions given in Section~\ref{sec:spatial-fields-resolution-sensitivity-analysis}.}
    \label{fig:hr-lr-sensitivity-avg-gains}
\end{figure}

Figure~\ref{fig:hr-lr-sensitivity-avg-gains} summarizes the visualizations of Figure~\ref{fig:lr-hr-sensitivity-analysis}.
We compare: (i) the difference between the relative gain brought by the in-situ time series (C1 against C3, the solid and dashed dark gray curves) and (ii) the difference between the relative gains brought by the spatial HR fields sampled at 12 hours against 24 hours (solid and dashed light gray curves). Referring to Figure~\ref{fig:lr-hr-sensitivity-analysis}, the dark gray solid line points represent the differences between dark and light green marked results, the dark gray dashed line points represent the differences between dark and light blue marked results. In the same way, the light gray solid point refer to the difference between the performances marked by dark green and blue bars and the points of the dashed light gray line refer to the differences between performances marked by light green and blue bars. For example, consider the point on the light gray curve in correspondence of the low-resolution combination group C. There, one may consider both the performance gain (w.r.t. the baseline of group C) of $M_m$-C3 and $M_m$-C1. The point of the light gray curve for the group C represents the difference between these two gains. Stated otherwise, the performance surplus imputable to the configuration C3 against the performance of configuration C1, which is the value added by in-situ time series. 

\section{Conclusions}\label{sec:discussion}
In this work we presented an observation system simulation experiment based on an hybrid data assimilation and deep learning framework. The objective of our analyses was to evaluate the impact and value of different input sources of information on sea-surface wind speed. Our investigation has both scientific and operational relevance. On the scientific side, our analyses prove the potential of a trainable approach for the simultaneous exploitation of diverse oceanic observations. This approach allows the model to ingest spatio-temporally heterogeneous data and to use this complementary information to reconstruct time series of spatial wind speed fields. The reconstruction performance associated with this explicit multi-modal strategy for the 4DVarNet ($M_m$) proves superior to the performance of the same 4DVarNet model not implementing the trainable multi-modal approach ($M_s$). We showed that this framework brings the model to automatically learn the correction of possibly biased input data, thanks to the high-level data representation encouraged by the multi-modal induced features extraction. On the operational side, beside proving the added value of in-situ observations,  our results on the buoys sensitivity analyses may open the road to future work devoted to the predictive learning-based optimization of in-situ sensors installation. The results presented in this work can not provide a direct answer to this problem but are an encouraging starting point to further investigate this aspect.

To conclude, we may highlight that the results we presented lack the fundamental aspect of uncertainty quantification. Recent work included this aspect by explicitly including an Evidence Lower BOund (ELBO) constraint in the 4DVarNet variational cost (see~\cite{Lafon2023}). This method allows to reconstruct the posterior distribution $p(\mathbf{x} | \mathbf{y})$ and not only the \emph{most likely} state, as done by the model described in this work. Likewise,~\cite{Beauchamp2023neural} propose an hybrid framework that leverages both the 4DVarNet scheme and stochastic partial differential equations to address jointly the interpolation and uncertainty quantification issues. We believe that our framework endowed by the capability of addressing the output distribution could provide insightful results on both the target variable reconstruction and  the quantification of its uncertainty.

\section*{Acknowledgement}
This work is funded by the AI Chair Oceanix (ANR grant ANR-19-CHIA-0016) and is supported by the industrial partnership with Naval Group. The computing resources were provided by the Région Bretagne, via the project CPER AIDA (2021 - 2027).

\newpage
\bibliographystyle{unsrtnat}
\bibliography{bibfile.bib}

\newpage
\appendix
\section{Choice of dynamical model parameterization}\label{app:app-model-phi}
In this appendix, we provide the following supplementary information. We argument the choice of the model architecture in Section~\ref{sec:app-model-sensitivity}. Sections~\ref{sec:app-models-recons} and~\ref{sec:app-lr-bias} present the different models reconstructions and the outcome of the LR bias analyses (cfr. Section~\ref{sec:lr-bias}) for different choices of the neural architecture used. We choose to perform this supplementary analysis to check if and in which measure a deeper and more complex architecture may impact the model capability to learn the correction for the biased LR data.

\subsection{Model complexity}\label{sec:app-model-sensitivity}
One primary point concerns the choice of the architecture of the model $\Phi$. This model is both used to parameterize the direct data-to-state variable inversion and to parameterize the dynamical operator of the 4DVarNet-based inversion scheme. Refer to Section~\ref{sec:methods} for an overview on the two approaches. 

\begin{figure}
    \centering
    \includegraphics[width=\textwidth]{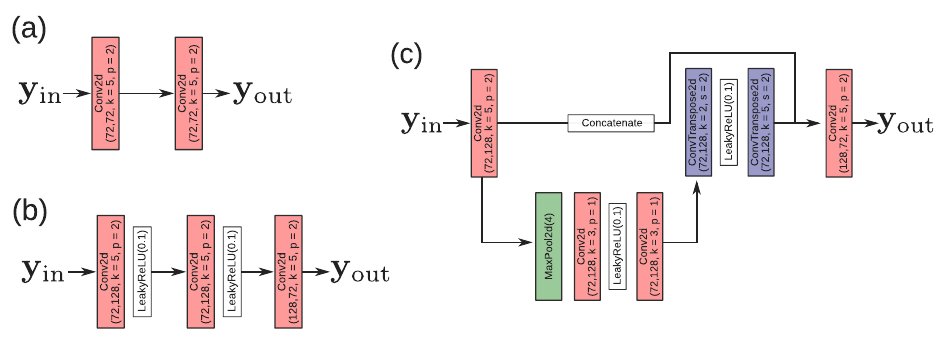}
    \caption{Sketch of the neural models tested. Panel (a): Model $\Phi_\alpha$. This is the model used for the analyses presented. Panel (b): Model $\Phi_\beta$. This model is similar to the simple network $\Phi_\alpha$ but is deeper, has more parameters and is non-linear. Panel (c): Model $\Phi_\gamma$. This architecture is a U-Net model. The red blocks represent Conv2d layers. Blue blocks represents Conv2d transposed layers. The symbols are the following. The first two argument of the Conv2d and Conv2d transposed functions are the input and output channels referred to one given layer. Letter \emph{k} is the kernel size, \emph{p} is the padding and \emph{s} is the stride. The green block represents a max-pooling 2d operation. The downsampling factor is $4$. The argument of the Leaky ReLU is the negative slope of the activation function.}
    \label{fig:appch4-models}
\end{figure}

The analyses presented in this work involve a simple neural architecture to parameterize the operator $\Phi$. Section~\ref{sec:numerical-implementation} described this architecture. The choice of such a simple model $\Phi$ is motivated by some preliminary sensitivity test w.r.t. model complexity. Let $\Phi_\alpha$ be the neural architecture described in Section~\ref{sec:numerical-implementation}. Let $\Phi_\beta$ and $\Phi_\gamma$ two more complex models. Figure~\ref{fig:appch4-models} illustrates these three models. $\Phi_\alpha$ and $\Phi_\beta$ are two convolutional networks. The first is a simple linear two-layered network while the second has one more layer and Leaky Rectified Linear Unit (ReLU) non-linear activation functions after the first two layers. The hidden channels dimension of $\Phi_\beta$ is $128$ while $\Phi_\alpha$ shrinks the channels dimension to $32$ in the connection between the two layers. 

The model $\Phi_\gamma$ is a U-Net. Contrarily to the first two models, $\Phi_\gamma$ implements downsampling and unsampling modules. After processing the input with a Conv2d layer, the downsampling module uses a max-pool and Conv2d layers to reduce the signal dimensions. The downsampled signal is processed by the upsampling module, which implements Conv2d transposed layers. The transposed convolutions are used to increase the dimension of the signal. This upampled signal is concatenated with the convolved input to create the U-Net shortcut. The signal is finally passed to a Conv2d layer.

\begin{table}[t]
    \centering
    \begin{tabular}{c c c c}
        \toprule[1pt]
         Model & RMSE [\si{\meter \per \second}] & Time & Memory size [\si{\mega \byte}] \\
         \midrule
         & \multicolumn{3}{c}{\textbf{Direct inversion}} \\
         $B_1(\Phi_\alpha)$ & $0.9571$ & $20$~min & $39$ \\
         $B_1(\Phi_\beta)$  & $0.9217$ & $30$~min & $129$ \\
         $B_1(\Phi_\gamma)$ & $0.8862$ & $40$~min & $430$ \\
         \midrule
         & \multicolumn{3}{c}{\textbf{4DVarNet}} \\
         $M_m(\Phi_\alpha)$ & $0.8617$ & $4$~hours~$40$~min & $157$ \\
         $M_m(\Phi_\beta)$  & $0.8604$ & $6$~hours~$40$~min & $248$ \\
         $M_m(\Phi_\gamma)$ & $0.8517$ & $7$~hours~$30$~min & $549$ \\
         \bottomrule[1pt]
    \end{tabular}
    \caption{Comparison of the three model architectures tested. The evaluation metrics reported in this table refer to 10 runs. The memory size is then the memory required to store the ensemble of the 10 models and the time is the training time of the 10 models ensemble. The models are trained under the C3 data configuration (cfr. Section~\ref{sec:exp-config}).}
    \label{tab:app-chap-2-models-comparison}
\end{table}

Table~\ref{tab:app-chap-2-models-comparison} reports the reconstruction error (the RMSE), the training times and the memory size for each one of the models introduced above. These metrics are referred to the ensemble of 10 models. So the RMSE is referred to the distance between the median aggregation and the ground truths (cfr. Section~\ref{sec:eval-framework}). The training time is the time required to train the ensemble of 10 models and the memory size is the memory required to save the ensemble of 10 models. Let
\begin{equation}
    \Delta E(\Phi) = \text{RMSE}_{B_1(\Phi)}(\hat{\mathbf{x}}_{\text{median}}, \mathbf{u}) - \text{RMSE}_{M_m(\Phi)}(\hat{\mathbf{x}}_{\text{median}}, \mathbf{u})
\end{equation}
be the difference between the RMSE of the direct inversion $B_1$ and the RMSE of the 4DVarNet $M_m$, as reported in Table~\ref{tab:app-chap-2-models-comparison}. The symbol $\hat{\textbf{x}}_{\text{median}}$ represents the aggregated output (cfr. Section~\ref{sec:eval-framework}) and $\mathbf{u}$ is the ground truth. The first remark is the following. The trend of the performance difference $\Delta E$ decreases with model complexity. The values for $\Delta E(\Phi_\alpha)$, $\Delta E(\Phi_\beta)$ and $\Delta E(\Phi_\gamma)$ are respectively $0.0954$, $0.0613$ and $0.0345$~\si{\meter\per\second}. This means that for the 4DVarNet inversion the performance improvement is more shrunken. Interestingly, the improvement is more evident for the case of direct inversion. In addition, the computational time required to train the 4DVarNet model is relevant. A gain of $10^{-3}$~\si{\meter\per\second} in terms of RMSE costs 2 hours more using $\Phi_\beta$ instead of $\Phi_\alpha$. Training the 4DVarNet model with $\Phi_\gamma$ as dynamical operator leads to a gain of $10^{-2}$~\si{\meter\per\second} in terms of RMSE w.r.t. the case of $\Phi_\alpha$ but for the price of nearly 3 more hours of training time. In light of these experiments, we choose to use the simple neural architecture $\Phi_\alpha$ to perform the numerical simulations. This choice is motivate by the extensiveness of the simulations required for the full analyses. However, we also think that an operationalization of such framework would require some further investigation on the optimal model architecture. 

\subsection{Reconstructions of the different models}\label{sec:app-models-recons}
\begin{figure}
    \centering
    \includegraphics[width=\textwidth]{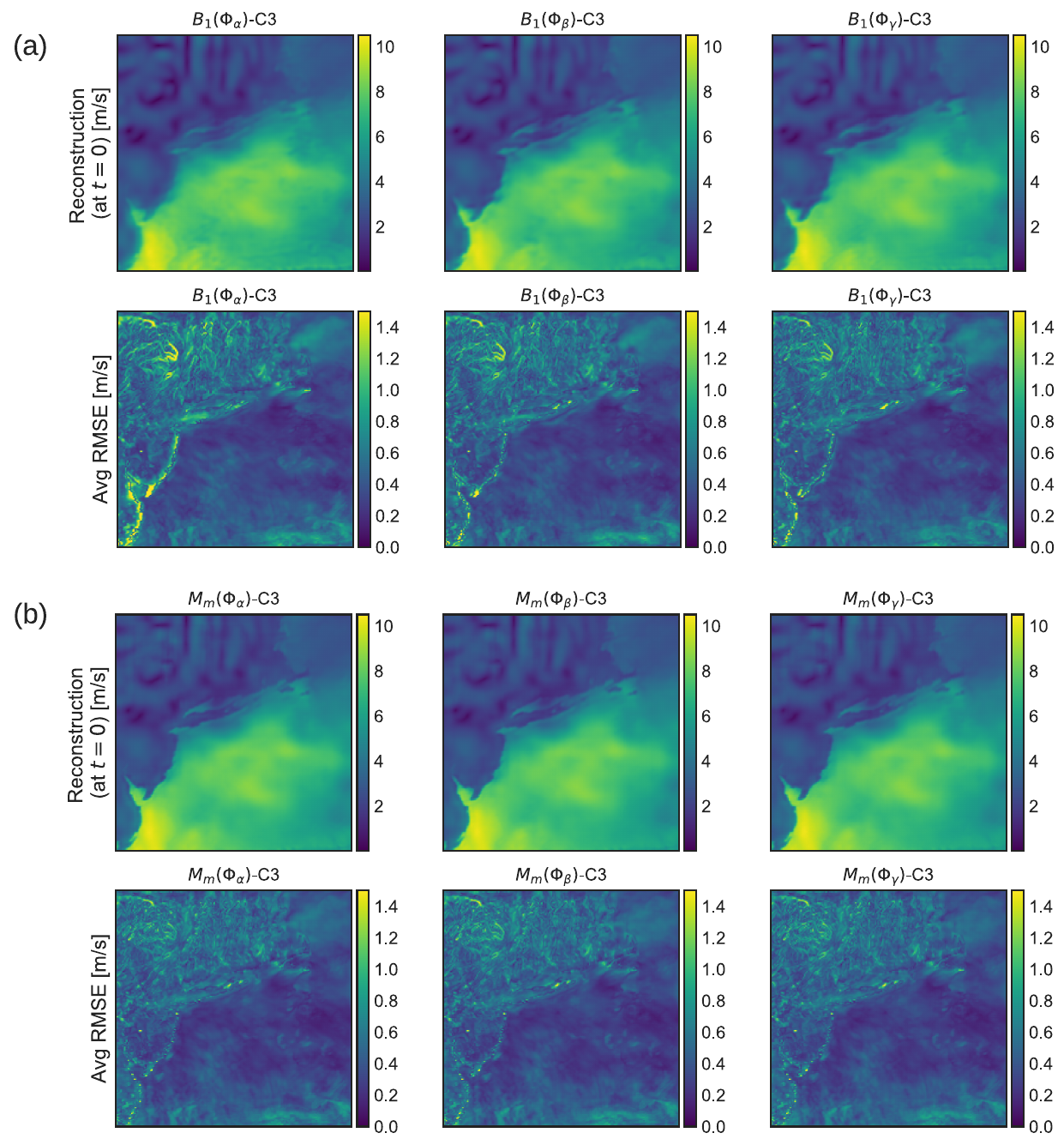}
    \caption{Reconstructions and RMSE visualizations for the three models discussed. Panel (a): Direct inversion. Panel (b): 4DVarNet inversion. In both panels, the top row depicts the reconstructions and the bottom row the RMSE of the model reconstructions w.r.t. the original data.}
    \label{fig:model-phi-alpha-beta-gamma}
\end{figure}

Figure~\ref{fig:model-phi-alpha-beta-gamma} depicts graphically the remark above. Panel (a) and (b) illustrate reconstructions and associated average error maps for the direct and 4DVarNet inversions backed by the three models presented above. For the reconstructions of the direct inversion case the improvement can be appreciated by naked eye. For the different 4DVarNet reconstruction this improvement is not visible. Recalling the results in Table~\ref{tab:app-chap-2-models-comparison}, the difference in terms of RMSE between these reconstructions is of the order of $10^{-2}$~\si{\meter\per\second}.

\subsection{LR biased data sensitivity}\label{sec:app-lr-bias}
\begin{figure}
    \centering
    \includegraphics[width=\textwidth]{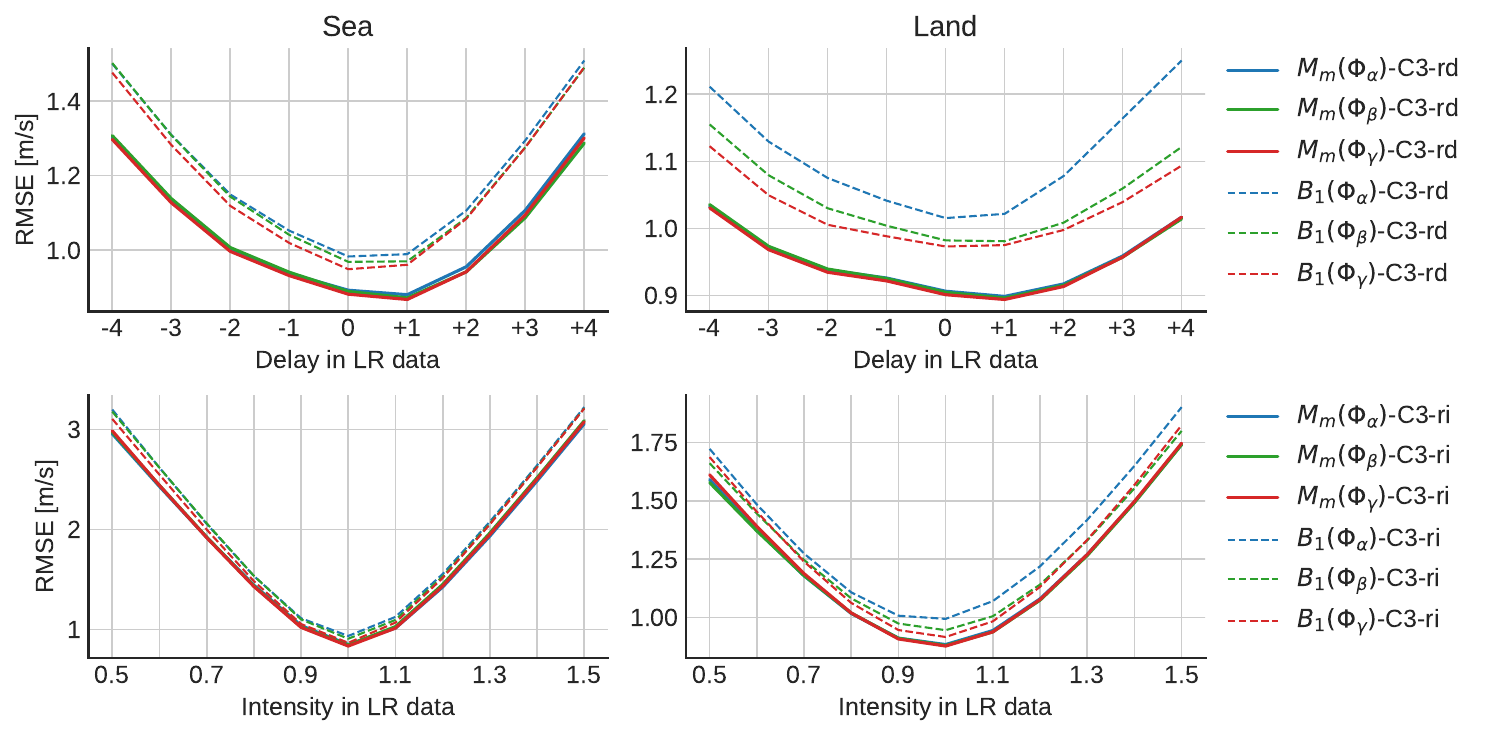}
    \caption{Degradation curves referred to the three models $\Phi_\alpha$, $\Phi_\beta$, $\Phi_\gamma$ for both direct inversion and 4DVarNet.}
    \label{fig:degradation-nets}
\end{figure}

\begin{figure}
    \centering
    \includegraphics[width=\textwidth]{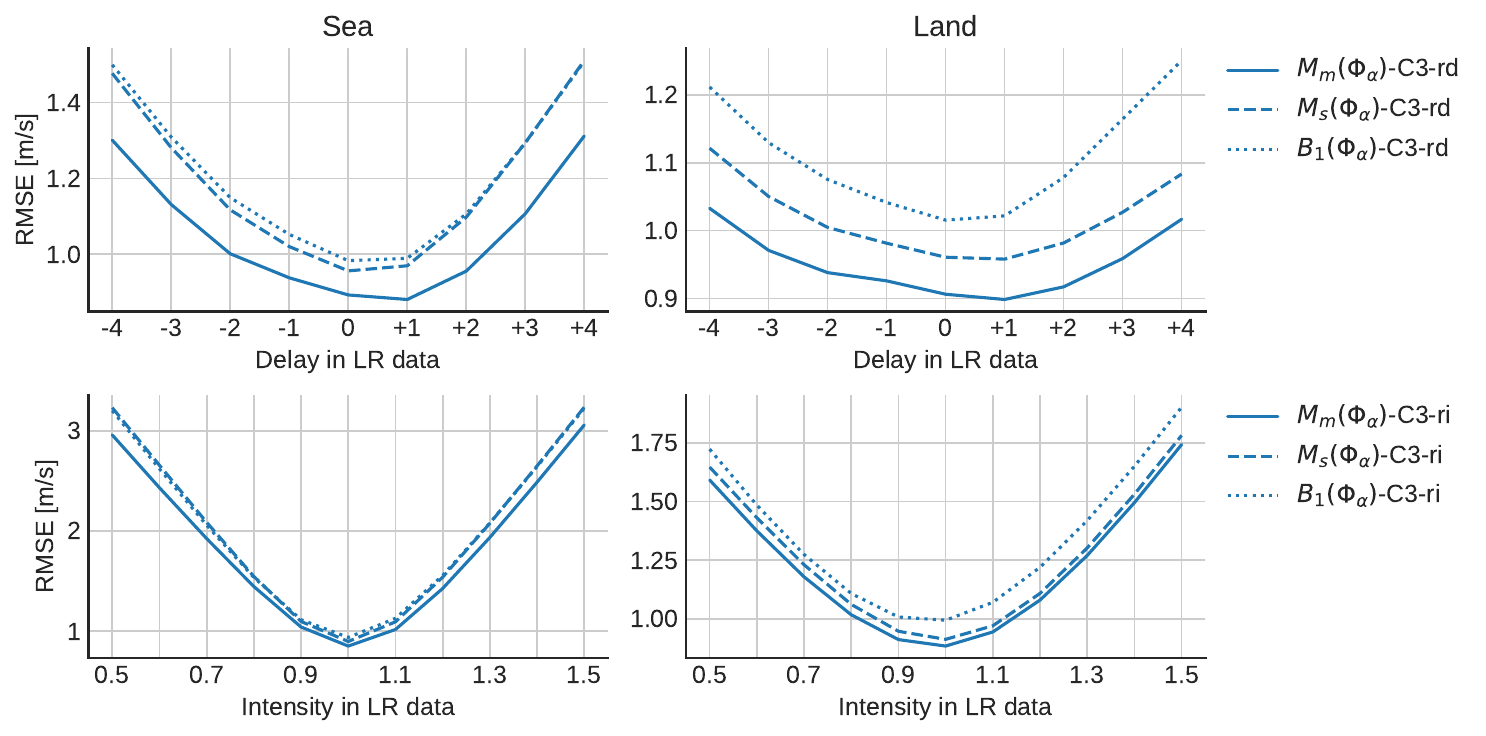}
    \caption{Degradation curves referred to $\Phi_\alpha$. This plot depicts the performance difference of direct inversion and the two instances of the 4DVarNet, that is with the simple and trainable data term in the variational cost.}
    \label{fig:degradation-nets-sn-sm-mm}
\end{figure}

The results on the biased LR data are reported in Figure~\ref{fig:degradation-nets}. These results corroborate the remarks provided above. The substantial differences between the model architectures are appreciable for the case of the direct inversion. In the case of the 4DVarNet framework such differences are minimal. This suggests that the overall 4DVarNet end-to-end architecture, constituted by the neural dynamical operator $\Phi$ and the gradient solver $\Gamma$ attenuate the limitations of a possibly under-sized and linear model like $\Phi_\alpha$. A further test, in Figure~\ref{fig:degradation-nets-sn-sm-mm}, aims to assess which is the difference between the performance of the two instances of the 4DVarNet model. This test is done for the model $\Phi_\alpha$. The difference between the curve related to the model $M_s$ (non-trainable data proximity term in the variational cost) and $M_m$ (trainable data proximity term) is evident. This may suggest that a relevant part of the 4DVarNet improvement and bias correction learning is implemented by the trainable multi-modal approach to process different data sources.

\end{document}